# Modeling Images using Transformed Indian Buffet Processes

**Yuening Hu**[†]    YNHU@CS.UMD.EDU
**Ke Zhai** [†]    ZHAIKE@CS.UMD.EDU
Department of Computer Science, University of Maryland, College Park, MD USA

**Sinead Williamson**    SINEAD@CS.CMU.EDU
Department of Machine Learning, Carnegie Mellon University, Pittsburgh, PA USA

**Jordan Boyd-Graber**    JBG@UMIACS.UMD.EDU
iSchool and UMIACS, University of Maryland, College Park, MD USA

## Abstract

Latent feature models are attractive for image modeling, since images generally contain multiple objects. However, many latent feature models ignore that objects can appear at different locations or require pre-segmentation of images. While the transformed Indian buffet process (tIBP) provides a method for modeling transformation-invariant features in unsegmented binary images, its current form is inappropriate for real images because of its computational cost and modeling assumptions. We combine the tIBP with likelihoods appropriate for real images and develop an efficient inference, using the cross-correlation between images and features, that is theoretically and empirically faster than existing inference techniques. Our method discovers reasonable components and achieve effective image reconstruction in natural images.

## 1. Introduction

Latent feature models assume data are generated by combining latent features shared across the dataset and aim to learn this latent structure in an unsupervised manner. Such models typically assume *all* properties of a feature are common to all data points—i.e., each feature appears in exactly the same way across all observations. This is often a reasonable assumption. For example, microarray data are designed so each cell consistently corresponds to a specific condition.

This does not hold for images. Consider a collection of images of a rolling ball. If a model must create new features to explain the ball's every position, it will devote less attention to other aspects of the image and will be unable to generalize across the ball's path. Instead, we would like *some* properties of a feature, e.g., shape, to be shared across data points but properties, e.g., location, to be observation-specific.

Models that generalize across images to discover transformation-invariant features have many applications. Image tracking, for instance, discovers mislaid bags or illegally stopped cars. Image reconstruction restores partially corrupted images. Movie compression recognizes recurring image patches and caches them across frames.

We argue that latent feature models of images should:

- **Discover features** needed to model data and add additional features to model new data.
- **Generalize across transformations** so features can have different locations, scales, and orientations.
- **Handle properties of real images** such as occlusion.

A nonparametric model that comes close to our goals is the noisy-OR transformed Indian buffet process (NO-tIBP, Austerweil & Griffiths, 2010); however, its likelihood model is inappropriate for real images. Existing unsupervised models that handle realistic likelihoods (Jojic & Frey, 2001; Titsias & Williams, 2006) are parametric and cannot discover new features. In Section 2, we further describe these and other models that meet some, but not all, of our criteria.

In Section 3, we propose models that fulfill these properties by combining realistic likelihoods with nonparametric frameworks. In Section 4, we introduce novel inference algorithms that dramatically improve inference for transformed IBPs in larger datasets (Section 5). In Section 6, we show that our models can discover features and model data better than existing models. We discuss relationships with other nonparametric models and extensions in Section 7.

† indicates equal contributions.

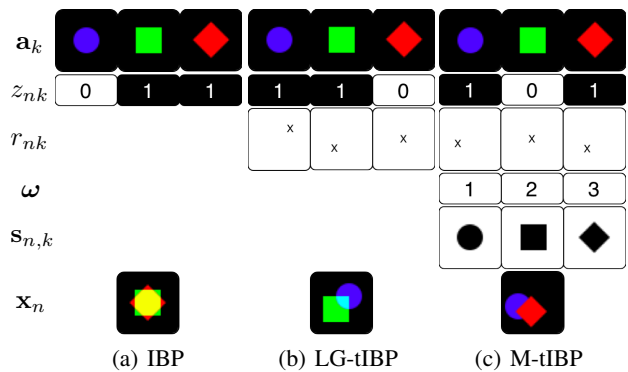

(a) IBP  (b) LG-tIBP  (c) M-tIBP

*Figure 1.* Generative process of the linear Gaussian IBP (IBP), the linear Gaussian tIBP (LG-tIBP) and the masked tIBP (M-tIBP). All models share features $\mathbf{a}_k$ across the dataset, and observation-specific indicators $z_{nk}$ determine which features contribute to a data point $\mathbf{x}_n$. In the tIBP models, transformations $r_{nk}$ change where features appear in the observation. In the IBP and LG-tIBP, features are combined additively. In the M-tIBP, only one feature contributes to each pixel of a final image. Together, a global ordering $\boldsymbol{\omega}$ over features and a local binary mask $\mathbf{s}_{n,k}$ determine which feature is responsible for the appearance of a given pixel.

## 2. Background

In this section, we review the Indian buffet process and how its extension, the transformed IBP, models simple images. We then describe likelihood models for images. These models are a prelude to the models we introduce in Section 3.

### 2.1. The Indian Buffet Process

The Indian buffet process (IBP, Griffiths & Ghahramani, 2005) is a distribution over binary matrices with exchangeable rows and infinitely many columns. This can define a nonparametric latent feature model with an unbounded number of features. This often matches our intuitions. We do not know how many latent features we expect to find in our data; neither do we expect to see all possible latent features in a given dataset.

To use the IBP to model data, we must select a likelihood model that determines the form of features corresponding to columns of $\mathbf{Z}$ and how the subset of features selected by a row of $\mathbf{Z}$ combine to generate a data point.[1] Many likelihoods have been proposed for the IBP, several of which are appropriate for modeling images.

### 2.2. The Transformed IBP

Most IBP-based latent feature models assume a feature is identical in every data point in which it appears. This precludes image modeling, where (for example) a car moves from location to location or where a person may be in either the foreground or background. Naïve models would learn different features for each location a car appears in; a more appropriate model would learn that each observation is in fact a transformation of a common feature.

The transformed IBP (tIBP, Austerweil & Griffiths, 2010) extends the IBP to accommodate data with varying locations. In the tIBP, each column of an IBP-distributed matrix $\mathbf{Z}$ is (as before) associated with a feature. In addition, each non-zero element of $\mathbf{Z}$ is associated with a transformation $r_{nk}$. Transforming the features and combining them according to a likelihood model produces observations. In the original tIBP paper, features were generated and combined using noisy-OR (Wood et al., 2006); we refer to this model as the noisy-OR tIBP (NO-tIBP), which allows the same feature to appear in different locations, scales, and orientations.

### 2.3. Likelihoods for Latent Feature Image Models

In addition to the noisy-OR, another likelihood that has been used with the IBP is a linear Gaussian model, which assumes images are generated via a linear superposition of features (Griffiths & Ghahramani, 2005). Each IBP row selects a subset of features and generates an observation by additively superimposing these features and adding Gaussian noise. This is demonstrated in Figure 1(a). This model can be extended by adding weights to the non-zero elements of the IBP-distributed matrix (Knowles & Ghahramani, 2007) and incorporating a spiky noise model (Zhou et al., 2011) appropriate for corrupted images.

If we want to model images where features can occlude each other, linear Gaussian models are inappropriate. In the vision community, images are often represented via overlapping layers (Wang & Adelson, 1994), including in generative probabilistic models (Jojic & Frey, 2001; Titsias & Williams, 2006). In these "sprite" models, features are Gaussian-distributed, and an ordering is defined over a set of features. In each image, every active feature has a transformation (as in the tIBP) and a binary mask for each pixel. Given the feature order, the image is generated by taking the value, at each pixel, of the uppermost unmasked feature.

This model is appealing. It is an intuitive occlusion model; features have a consistent ordering; and only the topmost feature is visible. However, this likelihood model has only been used for parametric feature sets and on data where the number of features is known *a priori*.

## 3. Modeling Real-valued Images

While the NO-tIBP likelihood model is incompatible with real images, it provides a foundation for nonparametric models with transformed features. In this section, we use the tIBP to build models that combine nonparametric feature

---
[1] We follow the convention that $\mathbf{z}_n$ is the $n^{th}$ row of a matrix $\mathbf{Z}$, and $z_{nk}$ is the $k^{th}$ element of the vector $\mathbf{z}_n$.

models with more useful and realistic likelihood functions for real images.

We begin by providing a general representation for the transformed IBP with an arbitrary likelihood.

1. Sample a binary matrix $\mathbf{Z} \sim \text{IBP}(\alpha)$, determining the features (columns) present in observations (rows).
2. For $k \in \mathbb{N}$, sample a feature $\phi_k \sim p(\phi)$.
3. For each image $n \in \{1, \ldots, N\}$
   - For $k \in \mathbb{N}$, sample a transformation $r_{nk} \sim p(r)$.
   - Sample an image $\mathbf{x}_n \sim p(\mathbf{x}|\mathbf{\Phi}, \mathbf{z}_n, \mathbf{r}_n)$.

The distribution over transformations $p(r)$, the feature likelihood $p(\phi)$, and the image likelihood $p(\mathbf{x}|\mathbf{\Phi}, \mathbf{z}_n, \mathbf{r}_n)$ can be defined in various ways. In the remainder of this section, we will use this generic framework to define concrete models with a parameterization of transformations and two different likelihood models.

**Transformations** Following Austerweil & Griffiths (2010), we consider three categories of transformation: translation, rotation and scaling. We parameterize a transformation $r : \mathbb{R}^D \to \mathbb{R}^D$ using a vector $(r_x, r_y, r_r, r_s)$. The parameters $(r_x, r_y)$ parameterize translations, and the transformed feature $r(\mathbf{a}_k)$ is obtained by shifting each pixel in $\mathbf{a}_k$ by $(r_x, r_y)$. Rotations are parameterized by $r_r \in [0, 2\pi)$, and scaling is parameterized by $r_s \in \mathbb{R}^+$. In practice, we restrict the possible rotations and scaling factors to a finite set, and assume a uniform prior on transformations.

**Linear Gaussian transformed IBP** Our first attempt to define a likelihood applicable to real data is based on the linear Gaussian likelihood for the IBP described in Section 2.3. Each feature $\phi_k$ is represented using a real-valued vector $\mathbf{a}_k \sim \mathcal{N}(\mathbf{0}, \sigma_a^2 \mathbf{I})$. In each image, the transformed features are combined using superposition,

$$\mathbf{x}_n \sim \mathcal{N}(\sum_{k=1}^{\infty} z_{nk} r_{nk}(\mathbf{a}_k), \sigma_x^2 \mathbf{I}). \quad (1)$$

We refer to the resulting model as the *linear Gaussian transformed IBP* (LG-tIBP).

**Masked transformed IBP** While the LG-tIBP model is appropriate for real-valued data, it cannot handle feature occlusion. To address this problem, we propose a *masked transformed IBP* (M-tIBP), based on the sprite model (Section 2.3). In this model, each feature $\phi_k$ is represented by a Gaussian feature $\mathbf{a}_k$ and a shape vector $\boldsymbol{\pi}_k$. Let $\boldsymbol{\omega}$ be a permutation of $\mathbb{N}$ that imposes an ordering on the features. We can interpret feature $i$ being "behind" feature $k$ if $\omega_i < \omega_k$. Each time a feature appears in an image, we sample a mask $\mathbf{s}_{n,k}$ from the Bernoulli probabilities in the corresponding shape vector $\boldsymbol{\pi}_k$. These masks "occlude" lower layers so that at each pixel; only the uppermost unmasked feature contributes to the final image.

The generative process can be described as follows. For each image $n$ and feature $k$, define an auxiliary variable $\mathbf{M}_{n,k}$, the visibility indicator,

$$M_{n,k}^d = \begin{cases} 1 & \text{if } \text{argmax}_j \left[\omega_j z_{n,j} \left(s_{n,j}^{r_{n,j}^{-1}(d)}\right)\right] = k \\ & \text{and } s_{n,k}^{r_{n,k}^{-1}(d)} > 0 \\ 0 & \text{otherwise}. \end{cases} \quad (2)$$

The visibility indicator $M_{n,k}^d$, is 1 when feature $k$ is the uppermost unmasked feature at pixel $d$ in image $n$. The image and feature likelihoods for the M-tIBP are

$$\begin{aligned}
\mathbf{a}_k &\sim \mathcal{N}(\mathbf{0}, \sigma_a^2 \mathbf{I}) \\
\pi_k^d &\sim \text{Beta}(\beta, \beta) \\
\omega &\sim \text{Uniform}() \\
\phi_k &:= (\mathbf{a}_k, \boldsymbol{\pi}_k, \omega_k) \\
s_{n,k}^d &\sim \text{Bernoulli}(\pi_k^d) \\
\mathbf{x}_n &\sim \mathcal{N}(\sum_{k=1}^{\infty} z_{nk} \cdot [r_{nk}(\mathbf{a}_k) \circ \mathbf{M}_{n,k}], \sigma_x^2 \mathbf{I}),
\end{aligned} \quad (3)$$

where the operator $\circ$ is the Hadamard product on matrices.

Figure 1 shows how the IBP-distributed matrix $\mathbf{Z}$ and other transformations variables combine features to form images for the IBP, LG-tIBP, and M-tIBP.

## 4. Inference

We perform inference of both LG-tIBP and M-tIBP using MCMC. At each iteration, we sample the Gaussian-distributed features $\mathbf{A}$, the IBP-distributed binary matrix $\mathbf{Z}$, the transformations $\mathbf{R}$, the hyperparameters $\alpha$, $\sigma_x$ and $\sigma_a$, and, for M-tIBP, the binary masks $\mathbf{S}$ and ordering $\boldsymbol{\omega}$.

### 4.1. Sampling Indicators, Transformations, and Masks

In all models, the binary indicator matrix $\mathbf{Z}$, the matrix of transformations $\mathbf{R}$, and (where appropriate) the feature masks $\mathbf{S}$ are all closely coupled. Austerweil & Griffiths (2010) sampled each $z_{nk}$ of $\mathbf{Z}$ by explicitly marginalizing over $r_{nk}$, and then sampling $r_{nk}$. However, explicitly computing the conditional distribution for all transformations for each feature cannot scale to even moderate-sized images (as discussed in Section 5). Instead, we sample $z_{nk}$, $r_{nk}$ and $\mathbf{s}_{n,k}$ jointly via a Metropolis-Hastings step.

The efficacy of a Metropolis-Hastings sampler depends on the quality of the proposal distribution. We design a data-driven proposal distribution (Tu & Zhu, 2002) $q(z_{nk}, r_{nk}, \mathbf{s}_{nk}) = q_z(z_{nk}) q_r(\mathbf{r}_{nk}) q_s(\mathbf{s}_{nk})$ based on an established pattern matching technique that assigns high probability to plausible states.

**Feature Indicator Proposal Distribution** Let $K_+$ be the highest feature index represented in the data, excluding the current data point. Our proposal distribution for $z_{nk}, k \leq K_+$ is

$$q(z_{nk} \to z_{nk}^*) = \begin{cases} 1 & \text{if } z_{nk}^* \neq z_{nk} \\ 0 & \text{otherwise.} \end{cases} \quad (4)$$

Our proposal distribution for previously unseen features follows Griffiths & Ghahramani (2005): sample $K^*$ new features according to Poisson$(\alpha/N)$, where $N$ is the number of observations.

**Transformation Proposal Distribution** To obtain a proposal distribution for translations $r_{nk}$ that matches our intuitions about the true posterior, we look at the *cross-correlation* between the feature $\mathbf{a}_k$ and the residual $\tilde{\mathbf{x}}_{n,k}$ obtained by removing all but that feature from the image $\mathbf{x}_n$. Cross-correlation (Duda & Hart, 1973) is a standard tool in classical image analysis and pattern-matching. The cross-correlation $\mathbf{u} \star \mathbf{v}$ between two real-valued images $\mathbf{u}$ and $\mathbf{v}$ is a measure of the similarity between $\mathbf{u}$ and a translated version of $\mathbf{v}$, i.e., $(\mathbf{u} \star \mathbf{v})(t) := \sum_{\tau=1}^T \mathbf{u}(\tau)\mathbf{v}(t+\tau)$.

Since our proposal distribution for $r_{nk}^*$ must be strictly positive, we use the exponentiated function

$$q(r|\mathbf{a}_k, \tilde{\mathbf{x}}_{n,k}) \propto \exp\{(\tilde{\mathbf{x}}_{n,k} \star \mathbf{a}_k)(r)\}, \quad (5)$$

for our proposal distribution,[2] and define the residual $\tilde{\mathbf{x}}_{n,k}$

$$\tilde{\mathbf{x}}_{n,k} = \sum_{j:\omega_j < \omega_k} M_{n,j} \circ \mathbf{x}_n \quad (6)$$

for M-tIBP, and

$$\tilde{\mathbf{x}}_{n,k} = \mathbf{x}_n - \sum_{j \neq k} z_{nj} r_{nj}(\mathbf{a}_j) \quad (7)$$

for LG-tIBP.

In Figure 2, we show the proposal distribution for $r_{nk}^*$ for a feature and three data points. The proposal distribution peaks in the locations that best match the pattern of pixels in the feature. If no locations match the feature, the proposal distribution is relatively entropic. Thus, the cross-correlation proposal distribution will cause us to consider good candidates for $r_{nk}$.

To incorporate scaling and rotation in addition to translation, we must increase the space over which we define our Metropolis-Hastings proposal. For a small transformation space (e.g., multiples of $\frac{\pi}{2}$ rotation and half / double scaling) it remains practical to extend the proposal distribution to include all possible scaling and rotation combinations. We separately obtain cross-correlations of these transformed features with the residual image, and concatenate the resulting vectors to obtain a distribution over all possible transformations. For new features, $r_{nk}$ is set to be the identity transformation.

---

[2] Of course, any $\mathbb{R} \to \mathbb{R}^+$ function would be a fair choice; however we found exponentiating works in practice.

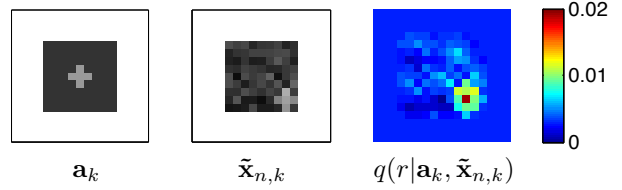

*Figure 2.* We use cross-correlation as our proposal distribution for the per-image, per-feature translation $r$. Here, we propose a translation $r_{nk}$ of a feature $\mathbf{a}_k$ that best explains the residual $\tilde{\mathbf{x}}_{n,k}$. Note that $r_{nk}$ need not lie within the boundaries of the image, so the borders for $\mathbf{a}_k$ and $\tilde{\mathbf{x}}_{n,k}$ indicate the range of possible $r_{nk}$.

**Mask Proposal Distribution** In the M-tIBP, we must also propose a binary mask $\mathbf{s}_{n,k}$. We use, as a proposal distribution, the conditional distribution

$$q_s(\mathbf{s}_{n,k}) = \prod_{d=1}^D p(s_{nk}^d = v | \mathbf{s}_{-(n,k)}) \quad (8)$$

$$p(s_{n,k}^d = 1 | \mathbf{s}_{-(n,k)}) = \frac{\sum_{m \neq n} s_{m,k}^d + \beta}{\sum_{m \neq n} z_{mk} + 2\beta}. \quad (9)$$

**Unseen Features** For previously unseen features, we sample a new feature $\mathbf{a}_k \sim \mathcal{N}(0, \sigma_a^2)$. Our proposal distribution for the corresponding mask is obtained by normalizing $\mathbf{a}_k$ and sampling each pixel of the proposed mask $\mathbf{s}_{n,k}^*$ according to a series of Bernoulli distributions parameterized by the normalized entries of $\mathbf{a}_k$.

### 4.2. Resampling Transformation and Masks

In addition to sampling $z_{nk}$, $r_{nk}$ and $\mathbf{s}_{n,k}$ jointly, we also resample $r_{nk}$ (and, for M-tIBP, $\mathbf{s}_{n,k}$) for values of $n$ and $k$ for which $z_{nk} = 1$. We jointly resample $r_{nk}$ using a Metropolis-Hastings step with proposal distribution $q_r(r_{nk})$ (or $q_r(r_{nk})q_s(\mathbf{s}_{nk})$). For the M-tIBP, we also Gibbs sample the binary masks using the conditional distribution

$$p(s_{n,k}^d | \mathbf{s}_{-(n,k)}^d, \mathbf{x}_n, \mathbf{z}, \mathbf{r}_n, \mathbf{A})$$
$$\propto p(x_n | \mathbf{s}_{n,k}^d, \mathbf{z}_n, \mathbf{r}_n, \mathbf{A}) \cdot p(s_{n,k}^d | \mathbf{s}_{-(n,k)}^d), \quad (10)$$

where $p(s_{n,k}^d | \mathbf{s}_{-(n,k)}^d)$ is given in Eqn. (9).

### 4.3. Sampling the Feature Order

We assume the feature order $\omega$ is sampled from a uniform distribution over permutations. We sample the feature order using a Metropolis-Hastings step where we uniformly choose two consecutive features and propose an order swap.

### 4.4. Sampling Features and Hyperparameters

Conjugacy eases the sampling of $\mathbf{a}_k$. For the M-tIBP, we sample the $d^{th}$ pixel of the $k^{th}$ feature as

$$a_{kd} | \mathbf{Z}, \mathbf{R}, \mathbf{S}, \mathbf{X} \sim \mathcal{N}\left(\tfrac{F}{\sigma_x^2} \sum_{n=1}^N M_{n,k}^d x_{n, r_{nk}(d)}, F\right), \quad (11)$$

where $F = (\sigma_a^{-2} + \sigma_x^{-2} \sum_{n=1}^{N} M_{n,k}^d)^{-1}$.

The hyperparameters $\alpha$, $\sigma_x$ and $\sigma_a$ can be Gibbs sampled via closed form equations (Doshi-Velez, 2009).

### 4.5. Modeling Color Images

The derivation above assumes that each pixel is a single real number. However, natural images are typically have color information, represented as a three-dimensional vector for each pixel. In our model, all colors contribute to the image likelihoods. Similarly, the proposal distribution is an element-wise sum over all possible channels,

$$q(r|\mathbf{a}_k, \tilde{\mathbf{x}}_{n,k}) \propto \exp\left\{\sum_c (\tilde{\mathbf{x}}_{n,k}^c \star \mathbf{a}_k^c)(r)\right\}, \quad (12)$$

where $\tilde{\mathbf{x}}_{n,k}^c$ and $\mathbf{a}_k^c$ are $c$-channel contribution of $\tilde{\mathbf{x}}_{n,k}$ and $\mathbf{a}_k$, respectively.

In the M-tIBP case, for feature $k$ in image $n$, we assume all channels share a common mask $\mathbf{s}_{n,k}$.

## 5. Computational Complexity

The main motivation behind the algorithm proposed in Section 4 is to allow the transformed IBP to be applied to large data. Austerweil & Griffiths (2010) calculate the likelihood of the data for every possible transformation. Replacing this naive approach with the sampler presented above can achieve a speed-up of at least $O(D\min(SR, K/\log D))$, where $R$ is the number of rotations considered, $S$ is the number of scales considered, $D$ is the number of pixels, and $K$ is the number of non-zero elements in $\mathbf{z}_n$.

Evaluating the LG-tIBP and M-tIBP likelihoods for a single image requires $O(DK)$ computations. Since the number of possible translations[3] is $O(D)$, calculating the likelihood for all possible translations in $O(SRD^2 K)$, yielding a total per-iteration complexity of $O(ND^2 K^2)$ for the inference method used by Austerweil & Griffiths (2010). If we were to also sum over values of $\mathbf{s}_{n,k}$, this would scale as $O(2^D)$.

By contrast, calculating the cross-correlation between a feature and an image residual can be done using the fast Fourier transform in $O(D \log D)$, so the proposal distribution described in Section 4.1 can be calculated in $O(SRD \log D)$. The likelihood need only be evaluated twice in the Metropolis-Hastings step, so our sampler scales as $O(NSRDK \max(K, \log D))$.

## 6. Experimental Evaluation

We evaluate the LG-tIBP and M-tIBP models[4] on both simulated and real-world data against the linear Gaussian IBP

---
[3]Since features can be centered outside the image, the total number of translations is in fact greater than the number of pixels.
[4]http://www.cs.umd.edu/~ynhu/code/mtibp

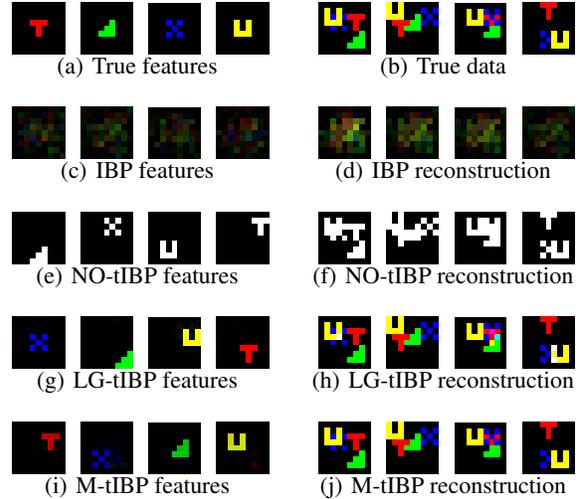

(a) True features    (b) True data
(c) IBP features    (d) IBP reconstruction
(e) NO-tIBP features    (f) NO-tIBP reconstruction
(g) LG-tIBP features    (h) LG-tIBP reconstruction
(i) M-tIBP features    (j) M-tIBP reconstruction

*Figure 3.* Comparing LG-tIBP and M-tIBP with NO-tIBP and IBP on synthetic data (image size $9 \times 9$) with **translation only**.

(IBP), the noisy-OR transformed IBP (NO-tIBP) and the sprite model (SPRITE, Jojic & Frey, 2001). Experiments on simulated data show that both LG-tIBP and M-tIBP recover the underlying features and locations more effectively than IBP. All data sets were scaled to have zero mean and unit variance for linear Gaussian models.

**Simulated Data** To qualitatively assess the ability of LG-tIBP and M-tIBP to find translated features, we generated data using four colorful features: "▽", "⊤", "⊔", and "×". Each synthetic dataset contains 100 images generated by selecting features independently with probability 0.5 and sampling a transformation uniformly. Since the noisy-OR likelihood cannot process color images, data are binarized for NO-tIBP. Although the other models can cope with Gaussian noise, NO-tIBP cannot, so no noise was added. Each experiment ran 100 iterations; we present features and reconstructions from the final iteration.

Figure 3 compares the performance of the four models on a dataset constructed by **translating** four features. NO-tIBP achieves good results. While the IBP struggles to find common structure, both LG-tIBP and M-tIBP generalize across locations and discover features qualitatively similar to NO-tIBP's. Where features overlap, M-tIBP obtains the correct reconstruction; LG-tIBP does not.

Figure 4 shows the training set likelihood at each iteration, plotted against accumulated CPU time, obtained using both the proposed Metropolis Hastings inference and Gibbs sampling in the LG-tIBP model on two datasets: $9 \times 9$ and $15 \times 15$ pixel images respectively. Each marker indicates a single iteration; each plot shows 100 iterations. Time was measured on a machine with 6-Core 2.8-GHz CPU and 16GB memory. The speed-up predicted in Section 5 is real-

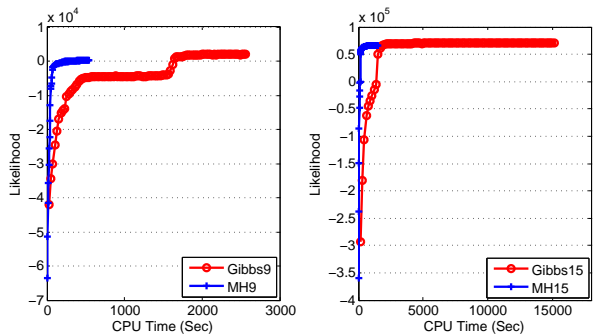

*Figure 4.* A run-time and likelihood comparison of using a Metropolis Hastings sampling of R and Z vs. Gibbs Sampling of R and Z. MH9 and Gibbs9 used a synthetic dataset with 9 by 9 pixel images, while MH15 and Gibbs15 used 15 by 15 pixels.

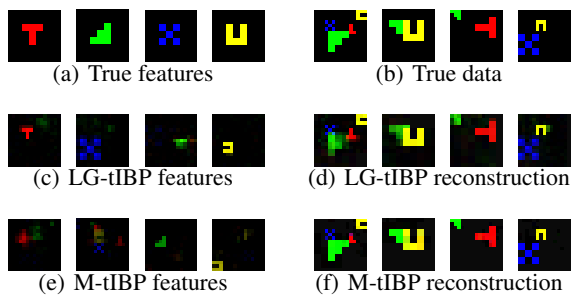

(a) True features  (b) True data
(c) LG-tIBP features  (d) LG-tIBP reconstruction
(e) M-tIBP features  (f) M-tIBP reconstruction

*Figure 5.* Evaluation of LG-tIBP and M-tIBP on synthetic data (image size $15 \times 15$) with **translation, rotation and scaling**.

ized in practice; while convergense requires slightly more iterations, it requires far less total CPU time.

In addition, we trained LG-tIBP and M-tIBP on a dataset where features have been **scaled, rotated, and translated**. This was not implemented by Austerweil & Griffiths (2010), presumably due to the computational cost. Figure 5 shows that our two models successfully detected the underlying features. The ordering learned by M-tIBP matches the true order, except in the case of the green "$\nabla$" and the blue "$\times$", which did not often overlap.

**Real-world data**  To show that the performance on simulated data in Section 6 carries over to real images, we evaluated LG-tIBP and M-tIBP on four image datasets, chosen to reflect various levels of complexity from simple video games with static/dynamic background to real-world scenes.

1. DNK: 171 screen shots from the 1981 video game "Donkey Kong".[5]
2. SMB: 200 screen shots from the 1985 video game "Super Mario Brothers".[6]
3. TFC: 186 frames from an intersection traffic video.[7]

[5] http://www.youtube.com/watch?v=EhFV5-qbbIw
[6] http://www.youtube.com/watch?v=xkD7L2QFwR0
[7] Raw AVSS PV Easy data available at http://www.eecs.qmul.ac.uk/~andrea/avss2007_d.html

|  | IBP | SPRITE | LG-tIBP | M-tIBP |
|---|---|---|---|---|
| DNK | 0.098 | 0.093 | 0.064 | 0.079 |
| SMB | 0.144 | 0.202 | 0.078 | 0.045 |
| TFC | 0.131 | 0.070 | 0.083 | 0.084 |
| WLK | 0.154 | 0.059 | 0.081 | 0.067 |

*Table 1.* Test set per-pixel per-channel RMSE (lower is better) on four datasets. The number of features for SPRITE is set to be the true number of features of each dataset. LG-tIBP and M-tIBP outperforms IBP on all datasets. M-tIBP works better than LG-tIBP on SMB and WLK, equally well with LG-tIBP on TFC, worse on DNK. M-tIBP performs equally well with SPRITE on three datasets, and outperforms SPRITE on SMB.

4. WLK: 226 frames from a video of people walking in a Lisbon shopping center.[8]

All images were resized to $101 \times 101$ pixels. We trained and tested the models using the full three-channel RGB data.

For each dataset, we trained LG-tIBP, M-tIBP, IBP and SPRITE[9] on a randomly selected $80\%$ of the images with the remaining $20\%$ held out for testing. Since the NO-tIBP is only appropriate for binary data, we could not compare with this method. We used the features extracted from the training set to estimate **Z** and **R** on test data, and evaluated the reconstructions using test set RMSE. Table 1 shows that LG-tIBP and M-tIBP achieve better performance than IBP across all datasets; M-tIBP performs equally well as SPRITE on three datasets, and much better on the SMB dataset. M-tIBP performs better than LG-tIBP on SMB and WLK datasets, but worse than LG-tIBP on DNK. This is because DNK has limited occlusions and a black background, and so can be adequately represented using the simpler LG-tIBP.

Figure 6 shows reconstructions and features obtained using the IBP, SPRITE, LG-tIBP, and M-tIBP. The IBP only matches the image background. In contrast, both LG-tIBP and M-tIBP identify shapes that appear in different locations. For example, in the first column of Figure 6, LG-tIBP identifies Donkey Kong (cyan) and a fireball (yellow), in addition to the background (green). Interestingly, LG-tIBP mis-identifies a pie[10] as a fireball but missed the actual fireball. Our M-tIBP model detected the pie (red) and the fireball (blue), while Donkey Kong (cyan) and background (yellow) are also clearly identified. Though M-tIBP has slightly larger RMSE than LG-tIBP on this dataset, the features seems more intuitive.

[8] Raw WalkByShop1cor data available at http://groups.inf.ed.ac.uk/vision/CAVIAR/CAVIARDATA1/
[9] The publicly available implementation of SPRITE could not detect any features in our datasets. To enable the fairest comparison possible, we compare against a finite version of M-tIBP with a fixed $K$ (based on the "true" $K$ based on inspecting the dataset, as in previous works using SPRITE) and an a "always on" $Z$. We believe that this is equivalent to the SPRITE *model*, although the *inference implementation* has tweaks and tricks that restrict the kinds of features that can be learned.
[10] "Pies" is the common name used for these sprites by Donkey Kong players; the designers' intent was to depict troughs of cement.

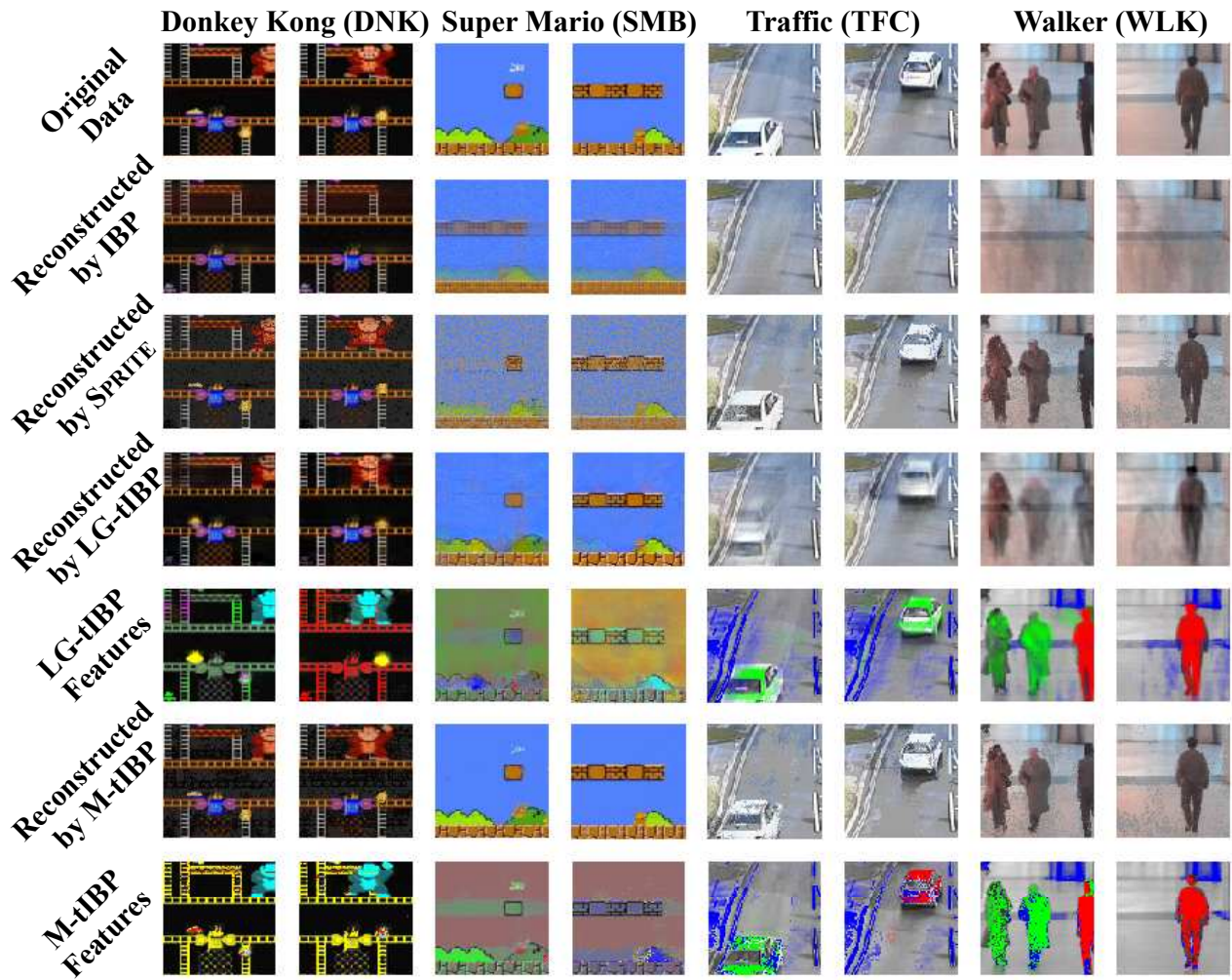

*Figure 6.* Reconstructions of test data by the IBP and LG-tIBP. First row: True image. Second, third, fourth and sixth rows: reconstructed image by IBP, SPRITE, LG-tIBP, and M-tIBP, respectively. Fifth and seventh rows: features detected by LG-tIBP and M-tIBP, respectively, superimposed on the true image. Each color is a feature; colors are consistent between columns. Each pair of adjacent columns are two images from the DNK, SMB, TFC and WLK datasets, respectively.

In the Super Mario dataset, while LG-tIBP extracted the bush and brick clearly, M-tIBP managed to extract the text "100", denoting points earned by the player (green). SPRITE performs poorly, possibly due to the large, sparsely observed feature set. M-tIBP identified the blue sky as two parts: one is the red feature and the other is the green feature. Because bricks often appear in the center of the screen, the model learns to "occlude" that location with a patch of sky.

While LG-tIBP and M-tIBP can learn features and transformations, M-tIBP is, on the whole, more accurate and the reconstructions are clearer. SPRITE can generally reconstruct data as well as M-tIBP, but the extracted features are less clear. One possible reason is that SPRITE assumes all features are present in each image. Moreover, in practice it is difficult to know *a priori* the number of features in a dataset. These two factors mean SPRITE is unlikely to scale to heterogeneous datasets such as SMB.

## 7. Discussion and Future Work

We have presented two nonparametric latent feature models for real-valued images, and presented a novel and efficient inference scheme. In this section, we discuss further applications of this inference paradigm, and discuss possible extensions to our models.

**Exploitation of Pattern Matching Algorithms** This inference scheme uses scoring functions from classical image analysis as the proposal distribution in a Metropolis-Hastings algorithm and combines the robustness and compu-tational appeal of a well-established pattern recognition tool with the flexibility of probabilistic models. This approach, or similar methods based on other classical pattern recognition techniques (Tu & Zhu, 2002; Tu et al., 2005), can be applied across a range of Bayesian models to improve inference in large state spaces.

An alternative to modeling images as real-valued vectors is to use image codewords (Li Fei-Fei & Perona, 2005). Other techniques have used transformed Bayesian nonparametric models to build high-performing vision systems using *fixed* codewords (Sudderth et al., 2005); a combination of these models would allow for a *joint* model to infer transformations, codewords, and feature cooccurrence patterns.

Rotation and scaling are implemented by extending the space for our cross-correlation-based proposal distribution. One avenue for future work is to investigate how existing non-statistical models for pattern recognition can sample a broader class of transformations using Metropolis-Hastings.

**Additional Modeling Directions** Features can appear more than once in an image, contrary to the assumptions of the tIBP. One avenue for future work is to extend the model to allow multiple instances of a feature in a given image. The infinite gamma-Poisson process (Titsias, 2007) is a distribution over infinite non-negative integer valued matrices. It has been used for image modeling, but that application required presegmentation of images. This work would allow extension to non-segmented images.

As in the original tIBP paper, we assumed that transformations associated with each (data point, feature) pair are sampled i.i.d. from some distribution $f(r)$ over possible transformations. One possible avenue for future research is to allow correlations (e.g., over time) between transformations in different images, leading to an image tracking model, which also leads to more efficient inference, by restricting the range in which an feature can appear in the $t^{th}$ to a neighborhood of the feature's location in the $(t-1)^{th}$ image. This idea was used to speed up inference in the SPRITE implementation of Titsias & Williams (2006). Incorporating spatial information into the mask distribution would also lead to more coherent feature appearances and counteract some of the "spotty" features observed for M-tIBP.

In addition, a more informative prior on the features $\mathbf{a}_k$ could be used to encode domain-specific knowledge—for example, for data comparable to the Walker video, one might make use of vertically oriented ellipses to find human-shaped features.

## 8. Acknowledgments

The authors would like to thank Joseph Austerweil, Frank Wood, Finale Doshi-Velez and Michalis K. Titsias for publishing implementations. This research was supported by NSF grant #1018625. Jordan Boyd-Graber is also supported by the Army Research Laboratory through ARL Cooperative Agreement W911NF-09-2-0072. Sinead Williamson is supported by NIH grant #R01GM087694 and AFOSR grant #FA9550010247. Any opinions, findings, conclusions, or recommendations expressed are the authors' and do not necessarily reflect those of the sponsors.